\documentclass[runningheads]{llncs}
\usepackage[T1]{fontenc}
\usepackage{subfiles}
\usepackage{amssymb}
\usepackage{comment}
\usepackage{microtype}
\usepackage{bm}
\usepackage{graphicx}
\usepackage{multirow}
\usepackage{xspace}
\usepackage{booktabs}
\usepackage{tabularx}
\usepackage{listings}
\usepackage{siunitx}
\usepackage{xcolor}
\usepackage{hyperref}
\usepackage{algpseudocode}
\usepackage{wrapfig}
\usepackage[most]{tcolorbox}
\newtcolorbox{todobox}[1]{
  enhanced,
  colback=blue!3!white,       
  colframe=blue!20!white,    
  coltitle=black!70,          
  fonttitle=\bfseries\sffamily,
  title={#1},
  boxrule=1pt,                
  arc=5pt,                   
  drop fuzzy shadow=black!5   
}

\lstdefinelanguage{SPARQL}{
  morekeywords={SELECT,WHERE,FILTER,BIND,OPTIONAL,ORDER,BY,ASC,DESC,GROUP,HAVING,FUSEJOIN,SIMJOIN，DEVJOIN},
  sensitive=true,
  morecomment=[l]{\#},
  morestring=[b]"
}

\lstdefinestyle{sparqlstyle}{
  language=SPARQL,
  basicstyle=\ttfamily\fontsize{7.8pt}{8.8pt}\selectfont, 
  keywordstyle=\bfseries,             
  numbers=left,                      
  numberstyle=\tiny,
  numbersep=5pt,
  frame=single,                      
  columns=fullflexible,
  keepspaces=true,
  showstringspaces=false,
  breaklines=true,
  aboveskip=0.5\baselineskip,         
  belowskip=0.5\baselineskip,
  abovecaptionskip=2pt,                
  belowcaptionskip=0pt,
  captionpos=b                        
}

\usepackage{amsmath}
\usepackage{enumitem}
\usepackage{xspace}
\usepackage{stmaryrd}

\newcommand\ma[1]{\ensuremath{\mathcal{#1}}}

\newcommand\tc[1]{\ensuremath{\textsc{#1}}}

\newcommand*\filter[2]{\ensuremath{\tc{Filter}_{\,#2}(#1)}}
\newcommand*\union[2]{\ensuremath{\tc{Union}(#1,#2)}}
\newcommand*\join[2]{\ensuremath{\tc{Join}(#1,#2)}}
\newcommand*\jointheta[2]{\ensuremath{\tc{Join}_{\Theta}(#1,#2)}}
\newcommand*\minus[2]{\ensuremath{\tc{Minus}(#1,#2)}}
\newcommand*\proj[2]{\ensuremath{\tc{Proj}_{\,#2}(#1)}}

\newcommand*\opt[3]{\ensuremath{\tc{Opt}_{\,#3}(#1,#2)}}
\newcommand*\diff[3]{\ensuremath{\tc{Diff}_{\,#3}(#1,#2)}}

\newcommand*\Dist{\ensuremath{\mathit{dist}}}
\newcommand*\eval[1]{\ensuremath{\llbracket #1 \rrbracket}}

\newcommand{\divtest}{\mathsf{prob{:}divergenceTest}}
\newcommand{\samedist}{\mathsf{prob{:}sameDistribution}}

\newcommand*\var[1]{\ensuremath{\operatorname{var}(#1)}}

\newcommand*\domain[1]{\ensuremath{\operatorname{dom}(#1)}\xspace}

\newcommand*{\GMM}{\mathsf{GMM}}

\newcommand{\Variables}{\ma V}
\newcommand{\BNodes}{\ma B}
\newcommand{\IRIs}{\ma I}
\newcommand{\Literals}{\ma L}
\newcommand{\Terms}{\ma T}
\newcommand{\Patterns}{\ma P}
\newcommand{\Filters}{\ma F}

\definecolor{probblue}{HTML}{0072B2}
\newcommand{\probext}[1]{\textcolor{probblue}{#1}}
\newcommand{\probfunc}[1]{\textcolor{probblue}{\texttt{#1}}}
\newcommand{\figlabel}[1]{\textsf{#1}}
\newcommand{\rdfterm}[1]{\texttt{#1}}
\begin{document}

\title{ProbSPARQL: Querying Knowledge Graphs with Multi-dimensional, Uncertain Numeric Data}

\titlerunning{ProbSPARQL: Querying KGs with Uncertain Numeric Data}

\author{Jingcheng Wu\inst{1}\thanks{Corresponding author.} \and
Ratan Bahadur Thapa\inst{1} \and
Daniel Hernandez\inst{1} \and
Hongkuan~Zhou\inst{1,2} \and
Steffen Staab\inst{1,3}}

\authorrunning{J. Wu et al.}

\institute{Institute for Artificial Intelligence, University of Stuttgart, Germany\\
\email{\{jingcheng.wu, ratan.thapa, daniel.hernandez, steffen.staab\}@ki.uni-stuttgart.de} \and
Bosch Corporate Research, Renningen, Germany\\
\email{hongkuan.zhou@de.bosch.com} \and
Web and Internet Science Research Group, University of Southampton, United~Kingdom}

\maketitle             
\begin{abstract}
The SFB 1574 ``Circular Factory'' is building a shared knowledge graph infrastructure for integrating data about returned products. A central challenge is that circular-factory data include numeric measurements that (i) originate from sensors or are derived from sensor-based measurements, (ii) are frequently multi-dimensional, and (iii) are inherently uncertain, while downstream triage, validation, reliability-modeling, and reassembly-planning modules require queryable uncertainty representations. Current RDF and SPARQL technologies lack native support for harmonized querying and analysis of such uncertain numeric measurement data. To address this gap, we present ProbSPARQL, an upward-compatible SPARQL extension developed as an early-stage query-layer pilot for this infrastructure. ProbSPARQL models uncertain numeric values as random variables whose distributions are encoded by probabilistic RDF literal datatypes, and supports distribution-aware expressions, probabilistic filters, and divergence-based joins. We implement ProbSPARQL on Apache Jena ARQ and expose it through a Fuseki-compatible execution layer. We assess real-data applicability using project-derived measurement fragments covering GMM-encoded uncertainty and histogram-based empirical roughness distributions, and evaluate scalability separately on controlled ontology-conformant benchmarks with up to 5,000 angle-grinder instances and 1.5M triples. The results show feasible in-engine execution, filter-pushdown speedups over application-layer post-processing, and latency--accuracy trade-offs among divergence-join decision strategies.
\end{abstract}

\section{Introduction}

Knowledge graphs~\cite{DBLP:series/synthesis/2021Hogan}, represented in RDF~\cite{world2014rdf} and queried via SPARQL~\cite{SPARQL11Query}, have been employed for data interoperability in manufacturing~\cite{DBLP:conf/kg4s/Thapa0BKHSSL25}, healthcare~\cite{rotmensch2017learning}, energy systems~\cite{DBLP:journals/tim/WangHG25}, and environmental monitoring~\cite{DBLP:journals/staeors/WuOOD24}. Most of these domains rely on physical sensors~\cite{DBLP:conf/cts/PatniHS10,DBLP:journals/ws/JanowiczHCPL19} that produce numeric data inherently subject to uncertainty~\cite{DBLP:journals/tkde/AggarwalY09,DBLP:conf/vldb/DeshpandeGMHH04}. 

This form of uncertainty appears concretely in the circular manufacturing scenarios studied in the Collaborative Research Centre SFB~1574 ``Circular Factory''~\cite{sfb1574,lanza2024vision}, a 12-year initiative funded by the German Research Foundation (DFG). The project aims to enable the perpetual use of technical products by integrating inspection, disassembly, reprocessing, and reassembly into a data-driven circular production system. Within this project, more than 60 researchers from engineering, computer science, and production-systems research across multiple institutions collaborate on a shared knowledge graph infrastructure for inspection and manufacturing data. The resulting infrastructure is therefore not merely a benchmark context, but a shared integration layer for engineering subprojects that produce, validate, and consume uncertain measurement data.

\begin{figure}[htbp]
    \centering
    \includegraphics[width=\linewidth]{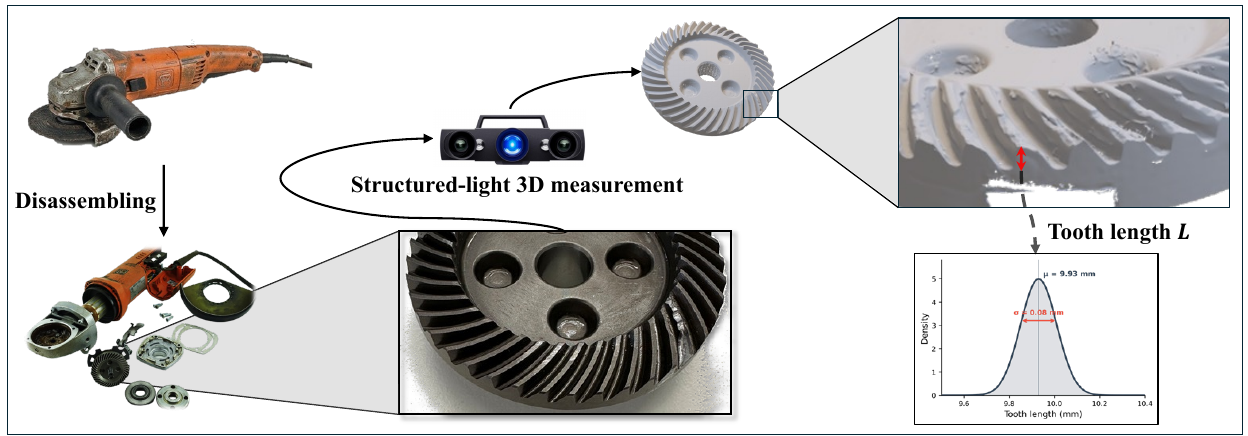}
    \caption{\small Circular manufacturing inspection workflow. An extracted crown gear undergoes Zeiss ATOS Q structured-light 3D scanning, yielding point clouds from which geometric features (e.g., tooth length~$L$) are modeled as probability distributions.}
    \label{fig:motivation}
\end{figure}

As illustrated in Figure~\ref{fig:motivation}, the inspection of returned industrial products is central to this goal. In the ``Circular Factory'' use case, a returned FEIN CG-15-125 BL angle grinder is disassembled, and its components undergo geometric inspection using a Zeiss ATOS Q structured-light 3D scanner. Extracting geometric characteristics such as individual tooth lengths from the resulting point clouds yields uncertain numeric data rather than exact scalar values: to account for sensor noise and reconstruction uncertainty, the geometric extraction pipeline outputs a probability distribution. These uncertain measurements must be represented in a common, machine-interpretable form while remaining queryable together with component identity, measurement context, and units in the shared knowledge graph infrastructure. Downstream triage, validation, reliability-modeling, and reassembly-planning modules require such queryable uncertainty representations. Analogous needs arise in healthcare and environmental monitoring, where measurement data likewise carry acquisition and device-specific uncertainty~\cite{rotmensch2017learning,DBLP:journals/staeors/WuOOD24}.

Yet, the current RDF and SPARQL specifications, comparable query languages (e.g., GQL \cite{ISO39075}), and specific ontologies such as the Semantic Sensor Network (SSN) Ontology \cite{w3c-ssn}, do not provide native support for harmonized querying and analysis of uncertain numeric measurement data. Such support requires that (i) multi-dimensional sensor measurement data is represented in an efficient manner, such as spatio-temporal point clouds; (ii) measurements are represented as random variables that define distributions over their data domains; (iii) essential probabilistic functions are natively supported; and (iv) selections, filters, and operations on random variables are integrated with both each other and other knowledge graph representations.

To address these requirements, this paper makes the following contributions:
\begin{enumerate}
    \item \textit{Probabilistic representation as random variables.} 
    We model uncertain numeric measurement as random variables defined by a domain and a distribution. By encoding such distributions as probabilistic RDF literals, we elevate them to first-class objects that can be accessed, queried, and used in computations.

    \item \textit{Distribution-aware query algebra.} 
    We extend the SPARQL algebra with a divergence join for distributional matching with a divergence tolerance and decision parameter, and introduce type-aware expression operators for comparing, transforming, fusing, and combining distributions.

    \item \textit{Extensible probabilistic datatype layer.} 
    We demonstrate the datatype layer with two parametric representations, Gaussian mixture models and Dirichlet distributions, and one non-parametric representation, histogram-based empirical distributions, all supported through a shared literal scheme and polymorphic operator interface.

    \item \textit{Implementation, evaluation, and artifacts.}
    We implement ProbSPARQL on top of Apache Jena and evaluate it through real-data applicability checks on project-derived circular-factory measurement fragments and controlled runtime experiments covering overhead, filter pushdown, divergence-join strategies, and datatype extensibility. We report early cross-institutional uptake within the SFB 1574 consortium, where non-author partner groups in inspection, materials analysis, and production-systems research provide consumer scenarios for downstream validation, matching, and planning modules, and release the implementation, generated evaluation data, query workloads, and replication scripts as reproducibility artifacts.\footnote{\url{https://github.com/0sidewalkenforcer0/ProbSPARQL}}
\end{enumerate}

\section{Use Case Requirements and Consuming Modules}
\label{sec:usecase}

The circular-factory scenario yields recurring query needs over uncertain
measurement data in the shared knowledge graph. The four use cases below are
representative rather than exhaustive and cover the current pilot's main
probability-aware query capabilities. They were formulated with engineering
partner groups in the cross-institutional SFB~1574 consortium based on recurring
data-access needs in inspection, materials analysis, and production-systems
research. These partner groups provide consumer scenarios for downstream
validation, matching, and planning modules beyond the ProbSPARQL proposer team.

\begin{small}
\begin{description}

    \item[R1: Threshold-based retrieval]
    (\textbf{Consumer:} Planning and control module.)\\
    \textsc{Purpose:} Automated triage agents screen returned gears for excessive tooth wear before routing them to reuse, reprocessing, or rejection. Since tooth-length measurements are uncertain, the module requires probabilistic rather than deterministic thresholding.\\
    \textsc{Query type:} Probabilistic thresholding (\textit{``Give me all gears which have at least one tooth whose length is below 9.8\,mm with probability at least 0.9.''})
    
    \item[R2: Anomaly detection]
    (\textbf{Consumer:} Autonomous measurement-strategy module.)\\
    \textsc{Purpose:} Measurement-validation components compare distributions obtained from different sensors or repeated measurements to detect sensor-specific anomalies, inconsistent reconstructions, or data-quality problems before the values are consumed by downstream modules.\\
    \textsc{Query type:} Distributional divergence (\textit{``Find all gears where the Jensen--Shannon divergence between the tooth-length distributions from CT and structured light exceeds 0.2.''})

    \item[R3: Motor performance evaluation]
    (\textbf{Consumer:} Reliability modeling module.)\\
    \textsc{Purpose:} Diagnostic agents evaluate the output power of returned motors for reliability modeling. Since power is not directly measured, it is derived from angular speed and torque, both represented as uncertain numeric values. The consuming module therefore needs a derived power distribution rather than a point estimate.\\
    \textsc{Query type:} Uncertainty propagation (\textit{``Retrieve all motors and compute their probabilistic power distributions derived from uncertain angular-speed and torque measurements.''})

    \item[R4: Divergence-based matching]
    (\textbf{Consumer:} Reassembly-planning module.)\\
    \textsc{Purpose:} Reassembly agents select compatible reclaimed drive and spindle gears for recombination. Substantially different hardness profiles may indicate contact-stiffness mismatch and vibration-relevant incompatibility.\\
    \textsc{Query type:} Divergence join (\textit{``Retrieve all available drive--spindle gear pairs for which the null hypothesis \(H_0 : \Delta \leq \epsilon\), where \(\Delta\) is the divergence between their hardness distributions, cannot be rejected at decision parameter \(\alpha = 0.05\).''})

\end{description}
\end{small}

\section{Core Concepts for Probabilistic Literals}
\label{sec:core-concepts}

Computing with probabilistic literals relies on modeling uncertain numeric values as \emph{random variables}. We classify \emph{operations on random variables} as distribution-valued transformations or deterministic-valued evaluations. A \emph{divergence test} is defined as a statistical hypothesis test.

\begin{definition}[Random Variable~\cite{rohatgi2015introduction}]
\label{def:rv}
Let $D \subseteq \mathbb{R}^d$ ($d \in \mathbb{N}^+$) be the \emph{domain}. A \emph{random variable} is a pair $X = (D, P)$, where $P$ is a probability measure over $D$. Two random variables $X = (D, P)$ and $X' = (D', P')$ are \emph{compatible} if and only if $D = D'$. Compatible random variables are \emph{distributionally equivalent}, written $X \equiv X'$, if and only if $P = P'$.
\end{definition}

\begin{definition}[Operations on Random Variables]
\label{def:operations}
Let $X_i=(D_i,P_i)$, for $i=1,\ldots,n$, be random variables with measurable domains $D_i \subseteq \mathbb{R}^{d_i}$. An operation on random variables is \emph{distribution-valued} if its result is a new random variable. It is \emph{deterministic-valued} if its result is a deterministic quantity computed from one or more underlying probability measures rather than a random variable.
\end{definition}

\begin{definition}[Divergence Test~\cite{lehmann2005testing}]
\label{def:div-test}
Let $X = (D, P)$ and $X' = (D, P')$ be compatible random variables, and let $\Delta$ be a statistical divergence measure satisfying $\Delta(X,X') = 0$ if and only if $X \equiv X'$. Given a tolerance $\epsilon \geq 0$ and a parameter $\alpha \in (0,1)$, the \emph{divergence test} formulates compatibility as the hypothesis-testing problem $H_0: \Delta(X, X') \leq \epsilon$ against $H_1: \Delta(X, X') > \epsilon$. The test regards $X$ and $X'$ as compatible when $H_0$ is not rejected by the configured decision strategy.
\end{definition}

Because $\Delta$ may lack a closed form, ProbSPARQL makes the decision strategy configurable and evaluates its latency--accuracy trade-off in Section~\ref{sec:evaluation}.

\section{ProbSPARQL Syntax and Datatype Interpretation}
\label{sec:probsparqlsyntax}

We adopt standard RDF/SPARQL notation~\cite{SPARQL11Query,world2014rdf}. 
Let $\Variables$, $\IRIs$, $\BNodes$, and $\Literals$ be pairwise disjoint sets of variables, IRIs, blank nodes, and literals, respectively, and let $\Terms = \IRIs \cup \BNodes \cup \Literals$ (as in~\cite{DBLP:journals/tods/PerezAG09}). 
We partition $\Literals$ into two disjoint subsets, $\Literals_{\mathit{rdf}}$ and $\Literals_{\Dist}$, where $\Literals_{\mathit{rdf}}$ are ordinary RDF literals and $\Literals_{\Dist}$ are RDF literals whose datatypes encode probability distributions.

\subsection{Representing Random Variables in RDF}
\label{sec:literals}

% We model a random variable as an RDF node that is linked both to its domain and to a probabilistic distribution encoded as a literal in $\Literals_{\Dist}$ (highlighted in green in Figure~\ref{fig:ontology-vocabulary}). This separation allows the RDF graph to represent the random variable explicitly as a distinct RDF node, while delegating the concrete encoding of the underlying probability distribution to a probabilistic literal.
We model a random variable as an RDF node linked to its domain and to a
probabilistic distribution literal in $\Literals_{\Dist}$ (green in
Figure~\ref{fig:ontology-vocabulary}). This keeps the random variable addressable in the RDF graph while delegating distribution parameters to a typed literal.

\begin{figure}[htbp]
  \centering
  \includegraphics[width=0.9\linewidth]{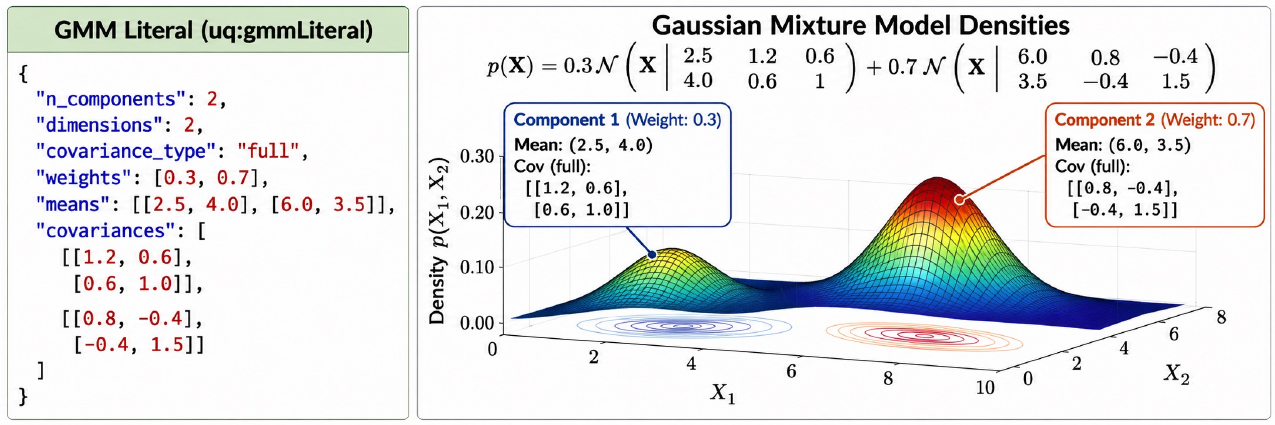}
  \caption{\small A 2D Gaussian mixture model instantiation. The left pane shows the JSON literal encoding the distribution parameters, while the right pane visualizes the corresponding probability density together with the mathematical form of the mixture.}
  \label{fig:gmm_2d_example}
\end{figure}

As a concrete instantiation, we introduce the \emph{Gaussian Mixture Model literal} (GMM literal) with the custom datatype IRI \texttt{uq:gmmLiteral}. Its value space is the set of Gaussian mixture models. Gaussian mixtures are widely used as a flexible approximation for continuous probability densities while still supporting tractable computations. Its lexical space $L_{\texttt{uq:gmmLiteral}}$ consists of JSON-encoded lexical forms~\cite{rfc8259}. Each lexical form must contain six mandatory fields: \texttt{n\_components}, \texttt{dimensions}, \texttt{covariance\_type} (\texttt{"full"}, \texttt{"diag"}, or \texttt{"spherical"}), \texttt{weights}, \texttt{means}, and \texttt{covariances}. Figure~\ref{fig:gmm_2d_example} illustrates a concrete two-dimensional GMM literal together with its JSON lexical form, corresponding density visualization, and mathematical form of the mixture.

\paragraph{Interpretation.} Each probabilistic datatype is associated with a lexical space, a value space,
and a partial interpretation from its lexical space to its value space, defined
only for datatype-specific well-formed lexical forms. For the datatype
\texttt{uq:gmmLiteral}, let $\GMM$ denote the value space of Gaussian mixture
models. The partial interpretation
$
\eval{\cdot}_{\Dist} : L_{\texttt{uq:gmmLiteral}} \rightharpoonup \GMM
$
maps a well-formed JSON lexical form to the corresponding GMM. If the JSON
lexical form is malformed or violates the required field-level constraints, the
interpretation is undefined. The formal GMM definition and well-formedness
constraints are provided in Section S.3.2 of the supplementary material~\cite{miscprobsparql2026}.

\paragraph{Extensible probabilistic datatypes.} The same datatype-interpretation scheme applies to other probabilistic datatypes: each datatype provides its own lexical syntax, well-formedness constraints, value space, and partial interpretation, while the RDF modeling pattern remains unchanged. 
Compositional data can be encoded using \texttt{uq:dirichletLiteral}, for example \texttt{\{"alphas": [5.0, 2.0, 3.0]\}}. 
Histogram literals can be encoded using \texttt{uq:histogramLiteral}, for example \mbox{\texttt{\{"dimensions": 1, "edges": [0, 10, 20], "weights": [0.3, 0.7]\}}}.

\subsection{Query Grammar and Expressions}
\label{sec:grammar}

ProbSPARQL follows the standard SPARQL graph-pattern syntax~\cite{DBLP:journals/tods/PerezAG09} and extends it with probabilistic filter functions and a $\theta$-join with \probext{\texttt{DIVJOIN}} syntax. Non-standard graph-pattern and filter constructs are highlighted in blue. A ProbSPARQL query is a graph pattern $\Patterns$ generated by
\begin{align*}
\Patterns\ &:= BGP \mid \filter{\Patterns}{\Filters} \mid \proj{\Patterns}{W}
          \mid \union{\Patterns}{\Patterns} \mid \join{\Patterns}{\Patterns} \\
          &\ \mid \probext{\jointheta{\Patterns}{\Patterns}}
          \mid \minus{\Patterns}{\Patterns}
          \mid \diff{\Patterns}{\Patterns}{\Filters}
          \mid \opt{\Patterns}{\Patterns}{\Filters},
\end{align*}
where $W \subseteq \Variables$ is the finite set of projected variables. Here, $\probext{\jointheta{\Patterns}{\Patterns}}$ denotes a parameterized join pattern with a finite set $\Theta$ of atomic divergence constraints. 
Each constraint is written abstractly as $?x \cong_{\epsilon,\alpha} ?y$, where $?x$ and $?y$ range over variables from the left and right $\Theta$-join operands, respectively, $\epsilon$ is the divergence tolerance, and $\alpha$ is the decision parameter of the corresponding test. 

Filters inherit the standard SPARQL filter language and are extended with concrete probabilistic filter functions. We use \probfunc{prob:divergenceTest}$(E,E,\epsilon,\alpha)$ for divergence testing and \probfunc{prob:sameDistribution}$(E,E)$ for exact distributional equality. For example, \texttt{FILTER(}\probfunc{prob:divergenceTest}\texttt{(?d1, ?d2, 0.1, 0.05))} invokes the divergence test with $\epsilon=0.1$ and $\alpha=0.05$. For the formal fragment considered, we restrict filters to safe expressions:
for every \(\filter{\Patterns}{\Filters}\), we require
\(\var{\Filters} \subseteq \var{\Patterns}\).

Following Definition~\ref{def:operations}, ProbSPARQL distinguishes two classes of probabilistic expressions: \emph{distribution-valued} expressions
$E^{\mathrm{dist}}$, which return random variables, and
\emph{deterministic-valued} expressions $E^{\mathrm{det}}$, which return
deterministic values. ProbSPARQL does not overload SPARQL arithmetic symbols
such as \(+\) and \(*\): arithmetic over scalar RDF literals remains unchanged, while arithmetic-like operations over distribution-valued literals are exposed through explicit distribution-valued operators with datatype-specific semantics.

\begin{definition}[Distribution-Valued Expression]
A distribution-valued expression $E^{\mathrm{dist}}$ is generated by
\[
\begin{aligned}
E^{\mathrm{dist}} ::=\, &
    ?x \mid c
    \mid \mathrm{scale}(E^{\mathrm{dist}}, \lambda)
    \mid \mathrm{shift}(E^{\mathrm{dist}}, v)
    \mid \mathrm{linearTransform}(E^{\mathrm{dist}}, A, b)
\\
& \mid \mathrm{marginal}(E^{\mathrm{dist}}, S)
    \mid \mathrm{joint}(E^{\mathrm{dist}}, E^{\mathrm{dist}})
    \mid \mathrm{mix}(E^{\mathrm{dist}}, E^{\mathrm{dist}}, w)
\\
& \mid \mathrm{multiply}(E^{\mathrm{dist}}, E^{\mathrm{dist}})
    \mid \mathrm{fuse}(E^{\mathrm{dist}}, E^{\mathrm{dist}})
    \mid \mathrm{convolve}(E^{\mathrm{dist}}, E^{\mathrm{dist}}),
\end{aligned}
\]
where $?x \in \Variables$ must be bound to probabilistic literals, and
$c \in \Literals_{\Dist}$.
\end{definition}

\begin{definition}[Deterministic-Valued Expression]
A deterministic-valued expression $E^{\mathrm{det}}$ is generated by
\[
\begin{aligned}
E^{\mathrm{det}} ::=\, &
    ?x \mid c
    \mid \mathrm{pdf}(E^{\mathrm{dist}}, t)
    \mid \mathrm{logpdf}(E^{\mathrm{dist}}, t)
    \mid \mathrm{cdf}(E^{\mathrm{dist}}, t)
    \mid \mathrm{mean}(E^{\mathrm{dist}})
\\
& \mid \mathrm{std}(E^{\mathrm{dist}})
    \mid \mathrm{quantile}(E^{\mathrm{dist}}, p)
    \mid \mathrm{map}(E^{\mathrm{dist}})
    \mid \mathrm{modeCount}(E^{\mathrm{dist}})
\\
& \mid \mathrm{jsd}(E^{\mathrm{dist}}, E^{\mathrm{dist}}),
\end{aligned}
\]
where $?x \in \Variables$ must be bound to ordinary RDF literals, and
$c \in \Literals_{\mathit{rdf}}$.
\end{definition}

The remaining symbols denote datatype-specific operator parameters.
Their signatures, result types, dimensionality requirements, and compatibility
conditions are given in Section S.4 of the supplementary material~\cite{miscprobsparql2026}.
\section{ProbSPARQL Query Semantics}
\label{sec:probsparqlsemantic}

We adopt the standard SPARQL algebra and evaluation framework~\cite{SPARQL11Query} and specify below only the modifications relevant to probabilistic literals and divergence joins.

\subsection{Distributional Compatibility}
\label{sec:algebra}

A \emph{solution mapping} is a partial function $\mu : \Variables \to \Terms$ with domain $\domain{\mu}$. The key modification to standard SPARQL lies in the treatment of equality on probabilistic literals. We define a binary relation $\equiv_{\Dist}$ on $\Terms$ by setting $t \equiv_{\Dist} t'$ iff either (i) $t,t' \in \Terms \setminus \Literals_{\Dist}$ and $t=t'$, or (ii) $t,t' \in \Literals_{\Dist}$ and $\eval{t}_{\Dist}=\eval{t'}_{\Dist}$. In all other cases, $t \not\equiv_{\Dist} t'$. Two solution mappings $\mu$ and $\mu'$ are compatible, written $\mu \sim \mu'$, iff $\forall ?x \in \domain{\mu}\cap\domain{\mu'}, \mu(?x)\equiv_{\Dist}\mu'(?x).$

\begin{definition}
If $\mu \sim \mu'$, then their union, written $\mu \uplus \mu'$, is the partial mapping over $\domain{\mu} \cup \domain{\mu'}$ defined by $(\mu \uplus \mu')(?x)=\mu(?x)$ for all $?x \in \domain{\mu}$ and $(\mu \uplus \mu')(?x)=\mu'(?x)$ for all $?x \in \domain{\mu'}$.
\end{definition}

Accordingly, ordinary join in ProbSPARQL follows the standard SPARQL join
structure, while compatibility on shared probabilistic literals is interpreted
via $\equiv_{\Dist}$ rather than by lexical identity of their encoded forms.\footnote{For example,
in a Gaussian mixture model with $k$ components, the syntactic ordering of the
components in the JSON encoding is irrelevant to the represented distribution.}

\subsection{Expression Evaluation}
\label{sec:expr-eval}

The evaluation of an expression $E$ under a solution mapping $\mu$, denoted $\eval{E}_{\mu}$, is defined recursively. Base cases are $\eval{?x}_{\mu}=\mu(?x)$ for variables $?x \in \Variables$ and $\eval{c}_{\mu}=c$ for literals $c \in \Literals$. For any $n$-ary built-in operator $\mathsf{op}$,
\[
\eval{\mathsf{op}(E_1,\dots,E_n)}_{\mu}
=
\textsc{Op}(\eval{E_1}_{\mu},\dots,\eval{E_n}_{\mu}),
\]
where $\textsc{Op}$ is the semantic function corresponding to $\mathsf{op}$. 
Detailed datatype-specific operator definitions are provided in Section S.4 of the supplementary material~\cite{miscprobsparql2026}. Probabilistic operators are partial and yield $\mathit{error}$ whenever their datatype-specific typing, lexical well-formedness, dimensionality, or compatibility requirements are not~satisfied.

\subsection{Probabilistic Filter Conditions}
\label{sec:prob-filters}

Standard SPARQL filter semantics are inherited, except that equality over probabilistic literals is interpreted via $\equiv_{\mathrm{dist}}$. Order comparisons such as $<$, $\leq$, $>$, and $\geq$ remain scalar: they are defined only for orderable RDF literals and yield $\mathit{error}$ on probabilistic literals. We specify below the additional probabilistic~filter~functions.

For a filter $\Filters$ and a solution mapping $\mu$, let
$\eval{\Filters}_{\mu}\in\{\top,\bot,\mathit{error}\}$ denote the truth-value of
$\Filters$ under $\mu$. Let $v_i=\eval{E_i}_{\mu}$ for $i=1,2$. When defined,
write $x_i=\eval{v_i}_{\Dist}$. Let $C_{\mu}$ hold iff $v_1$ and $v_2$ are
well-formed probabilistic literals and $x_1$ and $x_2$ are compatible. For
compatible $x_1$ and $x_2$, write $\mathrm{Acc}_{\epsilon,\alpha}(x_1,x_2)$ iff
the null hypothesis $H_0:\Delta(x_1,x_2)\le\epsilon$ cannot be rejected under the configured decision strategy with parameter $\alpha$. Then
\[
\eval{\divtest(E_1,E_2,\epsilon,\alpha)}_{\mu} =
\begin{cases}
\top, & \text{if } C_{\mu} \text{ and } \mathrm{Acc}_{\epsilon,\alpha}(x_1,x_2),\\
\bot, & \text{if } C_{\mu} \text{ and } \neg \mathrm{Acc}_{\epsilon,\alpha}(x_1,x_2),\\
\mathit{error}, & \text{otherwise}.
\end{cases}
\]
The filter function $\samedist$ provides an explicit form of distributional
equality:
\[
\eval{\samedist(E_1,E_2)}_{\mu} =
\begin{cases}
\top, & \text{if } C_{\mu} \text{ and } v_1 \equiv_{\Dist} v_2,\\
\bot, & \text{if } C_{\mu} \text{ and } v_1 \not\equiv_{\Dist} v_2,\\
\mathit{error}, & \text{otherwise}.
\end{cases}
\]

\subsection{$\Theta$-Join}
\label{sec:theta-join}

We next formalize the parameterized join $\Join_{\Theta}$. Let $\Theta$ be a finite set of atomic divergence constraints of the form $?x \cong_{\epsilon,\alpha} ?y$, where $?x$ and $?y$ are variables from the two join operands, $\epsilon$ is a divergence tolerance, and $\alpha$ is the decision parameter of the associated divergence test. For two solution mappings $\mu$ and $\mu'$, we write $\mu,\mu' \models (?x \cong_{\epsilon,\alpha} ?y)$ iff $?x \in \domain{\mu}$, $?y \in \domain{\mu'}$, $\mu(?x),\mu'(?y)\in\Literals_{\Dist}$, the interpreted probabilistic values $\eval{\mu(?x)}_{\Dist}$ and $\eval{\mu'(?y)}_{\Dist}$ are compatible, and the null hypothesis $H_0:\Delta(\eval{\mu(?x)}_{\Dist},\eval{\mu'(?y)}_{\Dist})\le\epsilon$ cannot be rejected at decision parameter $\alpha$. We write $\mu,\mu' \models \Theta$ iff every atomic constraint in $\Theta$ is satisfied. The $\Theta$-join of two sets of solution mappings $\Omega_1$ and $\Omega_2$ is then defined~by
\[
\Omega_1 \Join_{\Theta} \Omega_2
=
\left\{
\mu_1 \uplus \mu_2
\mid
\mu_1 \in \Omega_1,\ \mu_2 \in \Omega_2,\ \mu_1 \sim \mu_2,\ \mu_1,\mu_2 \models \Theta
\right\}.
\]

\begin{remark}
Unlike ordinary join, which enforces semantic equality on shared variables via $\sim$, a $\Theta$-join evaluates divergence constraints on designated variables from its two operands. For variable-disjoint operands, $\mu_1 \sim \mu_2$ holds vacuously, so the join is determined entirely by $\Theta$.
\end{remark}

\subsection{Evaluation Rules}
\label{sec:eval-rules}

The evaluation of a graph pattern $\Patterns$ over an RDF graph $G$, denoted $\eval{\Patterns}_G$, is defined as in standard SPARQL~\cite{SPARQL11Query}, except for the additional clause introduced by ProbSPARQL:
$
\eval{\jointheta{\Patterns_1}{\Patterns_2}}_G
=
\eval{\Patterns_1}_G \Join_{\Theta} \eval{\Patterns_2}_G.
$
The remaining recursive evaluation clauses, together with the algebraic characterization of the \(\Theta\)-join and its non-trivial optimization rules are given in Section S.9 of the supplement~\cite{miscprobsparql2026}.

% \subsection{Algebraic Characterization of $\Theta$-Join}

% The $\Theta$-join does not increase expressive power whenever its atomic divergence constraints are filter-definable. If the constraints in $\Theta$ can be represented by a filter formula $F_\Theta$, then a $\Theta$-join is equivalent to a standard join followed by selection:
% \[
% \Omega_1 \Join_\Theta \Omega_2 = \sigma_{F_\Theta}(\Omega_1 \Join \Omega_2).
% \]
% Thus, within the filter-definable fragment, $\Theta$-join is conservative over the ProbSPARQL algebra without the dedicated join operator. Its main role is to expose divergence-based matching as an optimizer-visible algebraic operator. The full propositions, proofs, and reassociation conditions are provided in the
% supplement.

\section{Circular Manufacturing Vocabulary}
\label{sec:vocabulary}
\begin{figure}[htbp]
    \centering
    \includegraphics[width=0.9\textwidth]{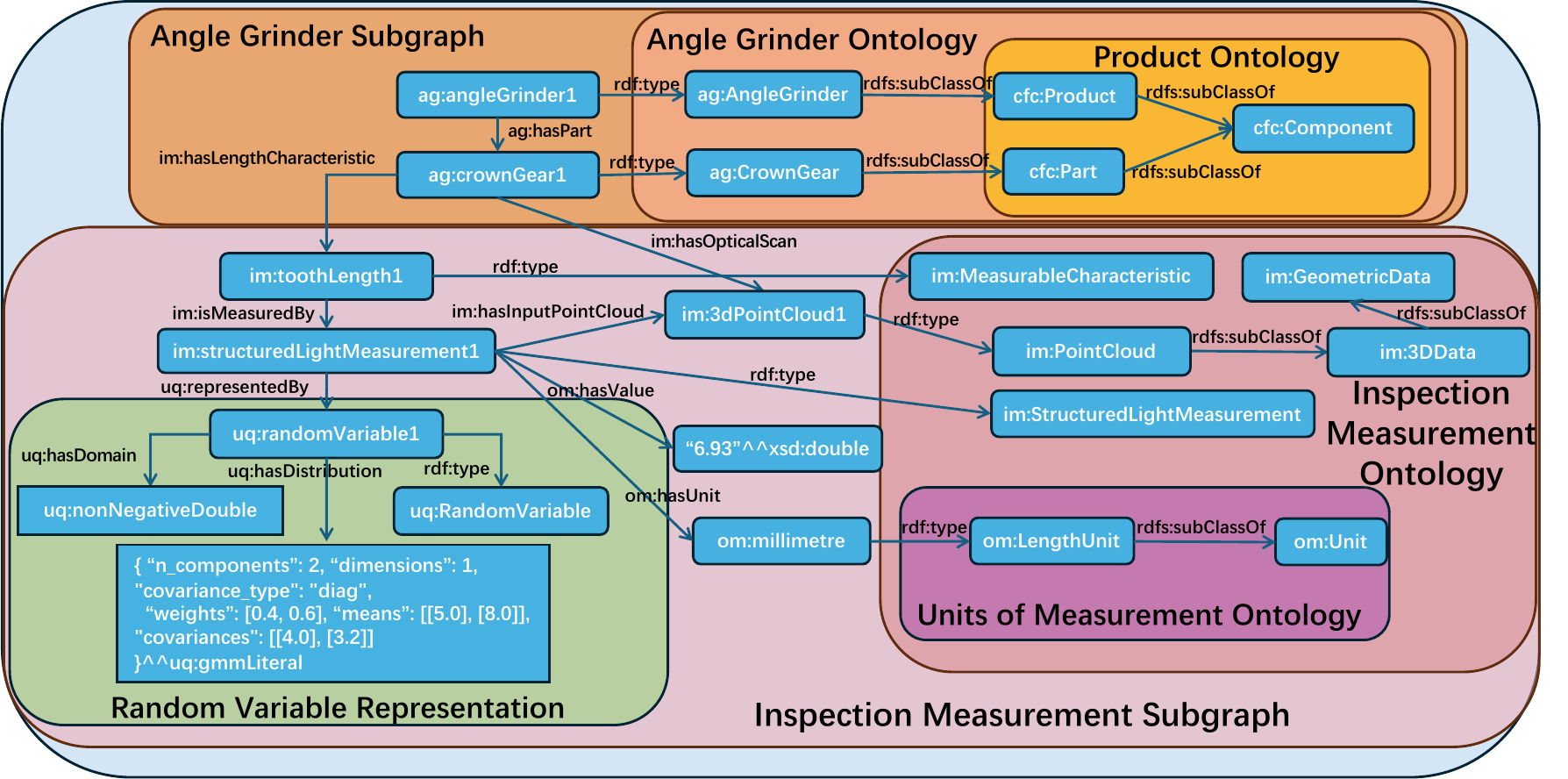}
    \caption{Circular manufacturing instantiation of the ProbSPARQL modeling pattern, connecting the \figlabel{Inspection Measurement Subgraph} to the \figlabel{Random Variable Representation} while preserving product structure, point-cloud input, scalar value, and unit information.}
    \label{fig:ontology-vocabulary}
\end{figure}

Figure~\ref{fig:ontology-vocabulary} shows the vocabulary pattern for the circular-factory KG. It combines the \figlabel{Angle Grinder Subgraph}, the \figlabel{Inspection Measurement Subgraph}, and the \figlabel{Random Variable Representation} with four ontology modules: the \figlabel{Product Ontology}\footnote{\url{https://w3id.org/circularfactory/PerpetualProduct}}, the \figlabel{Angle Grinder Ontology}, the \figlabel{Inspection Measurement Ontology}, and the \figlabel{Units of Measurement Ontology}. 
The \figlabel{Product Ontology} provides generic classes such as \rdfterm{cfc:Product}, \rdfterm{cfc:Part}, and \rdfterm{cfc:Component}, while the \figlabel{Angle Grinder Ontology} specializes them through \rdfterm{ag:AngleGrinder} and \rdfterm{ag:CrownGear}. The \figlabel{Inspection Measurement Ontology} captures concepts such as \rdfterm{im:MeasurableCharacteristic}, \rdfterm{im:PointCloud}, and \rdfterm{im:StructuredLightMeasurement}. The \figlabel{Units of Measurement Ontology}~\cite{omrij} supplies physical units such as \rdfterm{om:millimetre}.

The instance \rdfterm{im:structuredLightMeasurement1} in the \figlabel{Inspection Measurement Subgraph} is linked via \rdfterm{uq:representedBy} to the random-variable node \rdfterm{uq:randomVariable1} in the \figlabel{Random Variable Representation}, rather than being reduced to a detached uncertainty annotation. 
This node records a non-negative numeric domain via \rdfterm{uq:hasDomain} and a distribution encoded as a \rdfterm{uq:gmmLiteral} via \rdfterm{uq:hasDistribution}. 
This design keeps product structure, inspection provenance, scalar summaries, units, and uncertainty representation modular, while allowing graph patterns to retrieve the relevant distribution-valued literal for probabilistic evaluation.

\section{Implementation}
\label{sec:implementation}

Our open-source prototype extends the Apache Jena~\cite{apacheJena} ARQ query processor to evaluate probabilistic literals and ProbSPARQL operators during query execution.

The implementation introduces a polymorphic probabilistic datatype layer. Each supported datatype is associated with a parser from its lexical form to an internal value representation, a serializer back to RDF literals, and datatype-specific operator backends. For \texttt{uq:gmmLiteral}, JSON lexical forms are parsed into Gaussian mixture value objects used by backend routines for distribution evaluation, uncertainty propagation, and divergence estimation. Additional datatypes are supported by registering their datatype-specific parsers, serializers, value representations, and operator backends, including divergence or distance routines for comparison-oriented operators.

These extensions allow probabilistic expressions to be evaluated inside the SPARQL engine, rather than exporting distribution literals to an external application for post-processing. This in-engine evaluation reduces serialization and data-transfer overhead. The extended processor is deployed as an Apache Jena Fuseki endpoint and serves as the execution platform for the experimental evaluation in Section~\ref{sec:evaluation}.

\section{Evaluation}
\label{sec:evaluation}

We evaluate ProbSPARQL through four questions derived from the circular-factory query-layer requirements:
(EQ1) Can project-derived circular-factory measurement data be represented
and queried through the proposed random-variable and probabilistic-literal
pattern?
(EQ2) What overhead is introduced by probabilistic literals and operators
inside the SPARQL execution layer?
(EQ3) Does in-engine probabilistic filtering provide an integration benefit
over application-layer post-processing?
(EQ4) Which latency--accuracy trade-offs arise for divergence-based matching
in component-pairing workloads?

We answer EQ1 on a real circular-factory KG fragment derived from project
measurement data. We answer EQ2--EQ4 on controlled ontology-conformant
benchmarks. For mixture-complexity and operator-overhead experiments, we use
a fixed graph with 1,000 angle-grinder instances and approximately 300k triples,
varying the number of Gaussian mixture components \(K \in \{1,3,5,10\}\).
For graph-size scalability, we fix \(K=3\) and vary
the generated graph from 10 angle-grinder instances with 3,078 triples to 5,000
instances with 1,535,008 triples. The schema mirrors our circular-factory KG
infrastructure~\cite{lanza2024vision}, while the generated distributions provide
controlled workloads for comparing probabilistic query operators. All experiments were run on a consumer-grade laptop\footnote{Apple MacBook Pro
with M4 Pro chip, 48\,GB memory, macOS, Java~21, Jena~6.0.0.}. We report the
median of 10 runs after 3 warm-up iterations; standard deviations across runs
were below 5\% of the median for all measurements. Extended protocols, full result tables, complete query templates, workload-construction details, divergence-decision pseudocode, graph-size scalability, dimensionality scaling, and datatype-extensibility experiments are provided in Sections S.5–S.8 of the~supplement~\cite{miscprobsparql2026}.

We compare against baseline SPARQL execution whenever a deterministic
counterpart is available. We do not use existing probabilistic database systems
as runtime baselines because their data models and execution architectures are
not directly comparable to RDF graphs with distribution-valued probabilistic literals. Such a
comparison would conflate probabilistic-literal overhead with differences in data
model, encoding, and query execution. Moreover, these systems do not provide the
SPARQL-compatible divergence join required by R4.

\subsection{Pilot Validation on Real Circular-Factory Measurement Data}
\label{sec:eval-real}

To answer EQ1, we validate ProbSPARQL on a project-derived circular-factory
KG fragment. The graph stores material-analysis data for drive and spindle
gears, including Barkhausen-noise root-mean-square (RMS) signal measurements,
residual-stress measurements, hardness measurements, and process conditions.
Uncertain material-state measurements are represented as random variables with
GMM-encoded probabilistic RDF literals, while empirical roughness distributions
are represented through histogram literals. The graph contains 4,037 triples
across project-derived measurement subsets, with 97 sample resources,
20 process-condition resources, 280 measurement resources, and 163
random-variable resources. The main text shows R4 because it exercises the dedicated \texttt{DIVJOIN} operator. The query intent and representative bindings were reviewed with material-analysis and reassembly-planning partners outside the ProbSPARQL implementation team and matched the intended hardness-based scenario.

\begin{center}
\begin{minipage}{0.92\linewidth}
\begin{lstlisting}[style=sparqlstyle, caption={R4: Divergence join for component matching}, label={lst:R4-query}]
SELECT ?driveGear ?spindleGear WHERE {
  { ?driveGear a ag:DriveGear ; im:hasHardCharacteristic ?hardcharA .
    ?hardcharA im:isMeasuredBy ?measA .
    ?measA om:hasUnit om:HardnessVickers ; uq:representedBy ?rvA .
    ?rvA uq:hasDomain uq:nonNegativeDouble ; uq:hasDistribution ?distA . }
  DIVJOIN(?distA, ?distB, 0.3, 0.05)
  { ?spindleGear a ag:SpindleGear ; im:hasHardCharacteristic ?hardcharB .
    ?hardcharB im:isMeasuredBy ?measB .
    ?measB om:hasUnit om:HardnessVickers ; uq:representedBy ?rvB .
    ?rvB uq:hasDomain uq:nonNegativeDouble ; uq:hasDistribution ?distB . }
}
\end{lstlisting}
\end{minipage}
\end{center}

Listing~\ref{lst:R4-query} retrieves candidate drive--spindle gear pairs, follows the measurement links to
their hardness random variables, and extracts the corresponding probabilistic
hardness distributions. The call
\texttt{DIVJOIN(?distA, ?distB, 0.3, 0.05)} evaluates whether
\(H_0:\Delta \leq \epsilon\) cannot be rejected for the two distributions, using
\(\epsilon=0.3\) and configured decision parameter \(\alpha=0.05\). Matching is based on distributional compatibility rather than scalar equality.
This real-data execution demonstrates that project-derived hardness distributions
can be retrieved and matched directly by the probabilistic join operator. Runtime
behavior and scalability are evaluated next on the generated ontology-conformant
graphs.

\subsection{System Overhead and Filter Pushdown}
\label{sec:eval-overhead}

ProbSPARQL adds probabilistic literals and operators without changing the surrounding graph-pattern structure. We therefore isolate its latency overhead using deterministic counterparts for retrieval, threshold-based filtering (R1), and uncertainty-propagating arithmetic (R3). Distribution comparison (R2) is reported separately because it is dominated by Monte Carlo estimation and has no direct deterministic counterpart.

\begin{figure}[htbp]
  \centering
  \begin{minipage}[t]{0.48\textwidth}
    \centering
    \includegraphics[width=\linewidth]{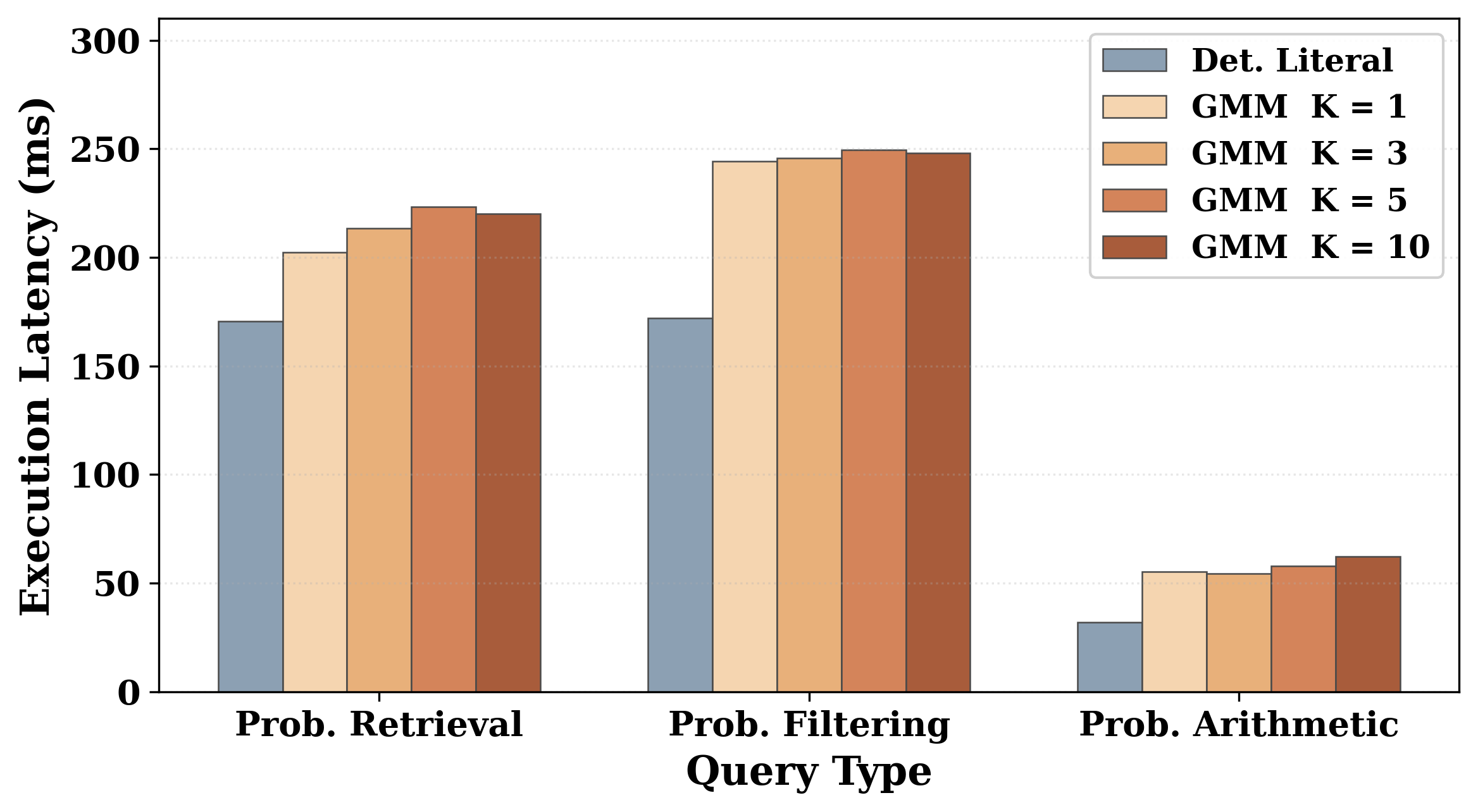}
    \caption{Execution latency for deterministic (Det.\ Literal) and probabilistic (GMM Literal) queries across mixture complexities $K \in \{1,3,5,10\}$. Distribution comparison is omitted due to its distinct sampling-based scaling behavior.}
    \label{fig:latency_overhead}
  \end{minipage}% 
  \hfill
  \begin{minipage}[t]{0.48\textwidth}
    \centering
    \includegraphics[width=\linewidth]{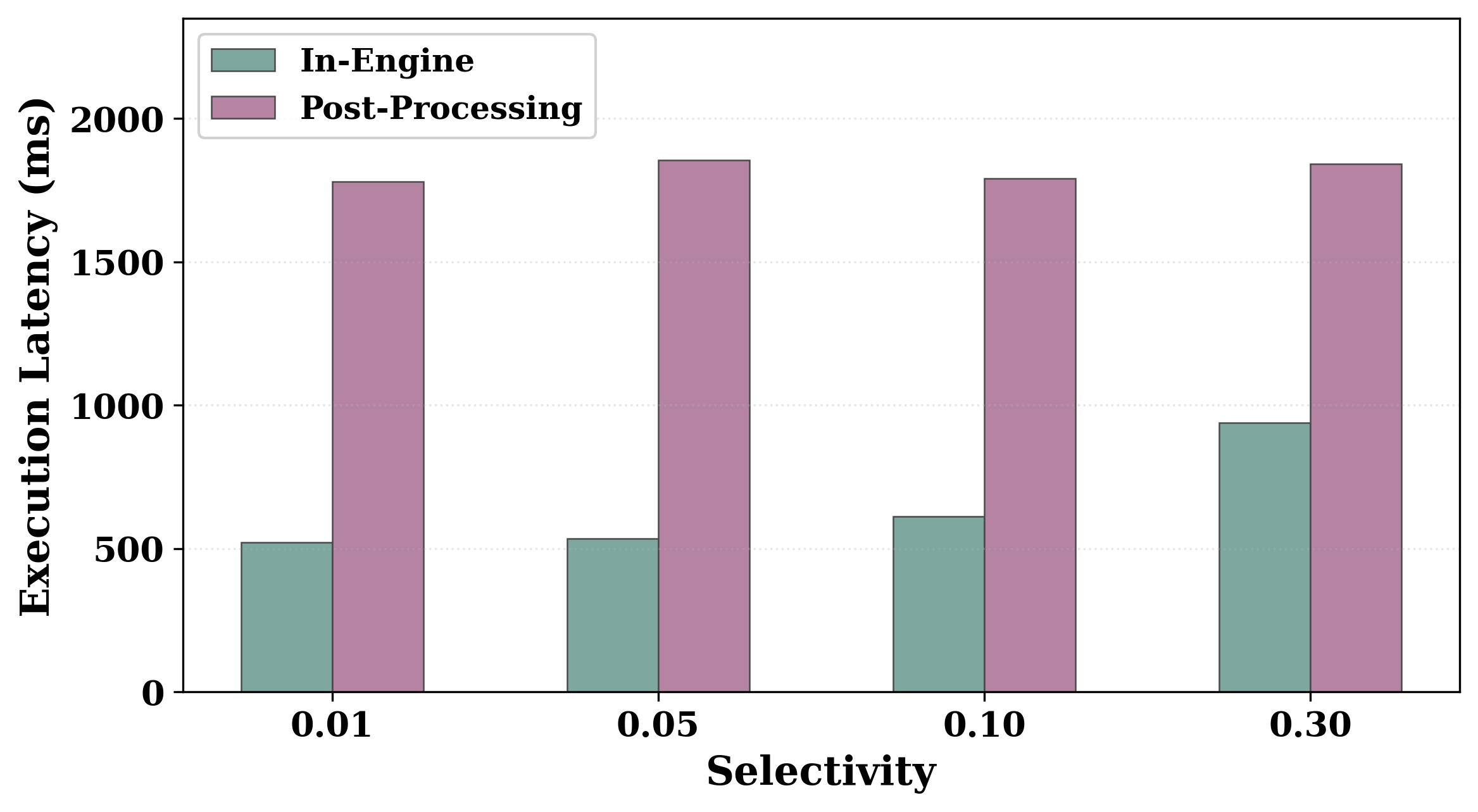}
    \caption{Latency comparison for CDF-based probabilistic filtering. In-engine evaluation avoids serialization and application-layer data transfer, achieving $2.0{\times}$--$3.5{\times}$ speedups depending on filter selectivity.}
    \label{fig:pushdown}
  \end{minipage}
\end{figure}

\smallskip\noindent\textbf{Execution overhead across mixture complexities.}
Figure~\ref{fig:latency_overhead} shows that retrieval incurs a $1.25$--$1.42{\times}$ overhead across all $K$ values, driven primarily by JSON literal deserialization. CDF-based filtering adds per-literal $O(K)$ CDF evaluations, raising the overhead slightly to $1.43$--$1.60{\times}$. Uncertainty-propagating arithmetic reaches $1.70$--$1.95{\times}$ due to the $O(K^2)$ closed-form moment computations in \texttt{prob:multiply}. Across all three query types, the weak dependence on $K$ indicates that parsing cost dominates the lightweight mathematical operations.

\smallskip\noindent\textbf{In-engine vs.\ post-processing performance.}
After quantifying the basic overhead of probabilistic literals and operators, we next measure the serialization and transfer costs avoided by in-engine
probabilistic filtering. The \emph{in-engine} strategy applies \texttt{FILTER(prob:cdf(?d, 9.8) >= 0.9)} before joining with downstream patterns. The \emph{post-processing} baseline retrieves all distributions and evaluates the CDF filter using Python's SciPy. Figure~\ref{fig:pushdown} reports median latencies across four filter selectivities spanning the range observed in our dataset, from highly selective (1\% of components flagged) to moderately selective (30\%). Post-processing latency remains at approximately 1\,800\,ms regardless of selectivity, as the entire set of GMM literals must be serialized and transferred over the network before filtering begins. In-engine evaluation applies the filter before serialization and transfer, achieving a $3.41{\times}$ speedup at a selectivity of 0.01, and maintaining a $1.96{\times}$ advantage even at 0.30.

The standalone distribution-comparison workload, driven by Monte Carlo sampling, scales from 1\,853\,ms ($K{=}1$) to 10\,744\,ms ($K{=}10$). This cost becomes critical in use case~R4, where divergence-test joins evaluate candidate pairs generated by the join operands. Section~\ref{sec:eval-simjoin} evaluates decision strategies that reduce the per-candidate-pair cost of divergence joins.

\subsection{Divergence Join Optimization}
\label{sec:eval-simjoin}

The divergence join evaluates the $\Theta$-join condition over candidate pairs
generated from the two join operands. In the prototype, this operator is exposed
through the \texttt{DIVJOIN} syntax shown in Listing~\ref{lst:R4-query}. While
the operator retains a quadratic worst-case pair enumeration, we exploit a key
algorithmic insight: evaluating $H_0:\Delta\leq\epsilon$ is a binary decision.
This avoids computing high-precision divergence estimates for every candidate
pair and enables substantial reductions in candidate-pair decision time via
bounding~\cite{DBLP:books/daglib/0016881} and early-stopping
mechanisms~\cite{sequentialtest}.

To evaluate these optimizations, we compare five decision strategies across controlled candidate-pair workloads with high-precision reference JSD values~\cite{endres2003new}, grouped by decision difficulty. The \emph{Hard} workload comprises near-threshold pairs whose reference JSD lies within $\epsilon \pm 0.05$, where estimation variance is likely to affect the binary decision. The \emph{Mixed} workload provides a balanced baseline by combining pairs from all difficulty levels in equal proportions. Table~\ref{tab:sampling} reports F1 and MAE over each workload, together with the median decision time per candidate pair.

\smallskip\noindent\textbf{Sampling strategies.}
The fixed-budget Monte Carlo baselines use $N=10{,}000$ samples per candidate pair. 
\emph{MC-Naive} applies standard Monte Carlo estimation. 
\emph{MC-Strat} uses variance-reduced stratified sampling~\cite{DBLP:books/sp/RobertC04}. 
\emph{SPRT} applies Wald's sequential probability ratio test~\cite{sequentialtest} for early termination in hypothesis testing, configured with significance level $\alpha=0.05$. 
\emph{Det-Bound} uses an analytic lower bound on JSD derived from the Data Processing Inequality~\cite{DBLP:books/daglib/0016881} as a deterministic rejection test. When used standalone, unresolved cases are treated as non-rejections of $H_0$. 
\emph{Adaptive-Cascade} first applies Det-Bound to reject $H_0$ when possible, then triggers SPRT for sequential evaluation, and finally falls back to MC-Strat for ambiguous boundary cases that exhaust the sample budget.

\begin{table*}[htbp]
  \centering
  \renewcommand{\arraystretch}{1.15}
  \caption{Divergence-test strategies on Hard and Mixed workloads. Results report F1 (\%), MAE, and median per-candidate-pair latency (ms).}
  \label{tab:sampling}
  \resizebox{0.88\textwidth}{!}{%
  \begin{tabular}{@{} l *{3}{>{\centering\arraybackslash}p{4.2em}} c *{3}{>{\centering\arraybackslash}p{4.2em}} @{}}
    \toprule 
    & \multicolumn{3}{c}{Hard Workload} & & \multicolumn{3}{c}{Mixed Workload} \\
    \cmidrule(lr){2-4} \cmidrule(l){6-8} 
    \textbf{Strategy} & \textbf{F1} & \textbf{MAE} & \textbf{Latency} & & \textbf{F1} & \textbf{MAE} & \textbf{Latency} \\
    \midrule 
    MC-Naive         & 96.7 & .0041 & 25.68 & & 99.3 & .0028 & 25.09 \\
    MC-Strat         & 97.7 & .0021 & 23.34 & & 99.7 & .0015 & 25.89 \\
    SPRT             & 94.3 & .0062 & 4.47  & & 98.0 & .0048 & 1.81  \\
    Det-Bound        & 94.9 & .0184 & 0.01  & & 98.0 & .0115 & 0.01  \\
    Adaptive-Cascade & 95.9 & .0107 & 3.27  & & 99.0 & .0078 & 1.04  \\
    \bottomrule 
  \end{tabular}%
}
\end{table*}

\smallskip\noindent\textbf{Results.}
Table~\ref{tab:sampling} compares five strategies across the two workloads. Since the framework is strategy-agnostic, users may select a strategy according to their latency and accuracy requirements. We use Adaptive-Cascade as the default strategy because it offers the most balanced performance across decision difficulties.

On the \emph{Hard} workload, MC-Strat achieves the highest F1 score (97.7\%) but requires 23.34\,ms per candidate-pair decision. Adaptive-Cascade lowers the decision latency to 3.27\,ms while maintaining a 95.9\% F1 score. This yields a $7.9{\times}$ speedup over MC-Naive and improves over standalone SPRT in F1 score (94.3\%). On the \emph{Mixed} workload, Adaptive-Cascade reaches 99.0\% F1 with a 1.04\,ms per-candidate-pair latency, yielding a $24.1{\times}$ speedup over MC-Naive.

% Although the standalone Det-Bound strategy offers near-zero latency (0.01\,ms), it achieves only 94.9\% F1 on the \emph{Hard} workload because the conservative lower bound is insufficient for many near-threshold decisions. Adaptive-Cascade addresses this limitation by using Det-Bound only for initial rejection of $H_0$, then applying SPRT, and finally falling back to MC-Strat when ambiguous boundary cases exhaust the sequential test. This raises the Hard-workload F1 score to 95.9\% while retaining low latency. A comprehensive precision/recall breakdown and end-to-end divergence-join evaluations are provided in the supplement.
Although Det-Bound is nearly cost-free, its conservative lower bound misses many
near-threshold decisions on the \emph{Hard} workload. Adaptive-Cascade therefore
uses it only as an initial rejection step and falls back to SPRT and MC-Strat for
unresolved cases, improving F1 to 95.9\% while retaining low latency.

% \Jingcheng{Add this to journal version}\\
% We organize related work into four areas: similarity-aware extensions of SPARQL, probabilistic relational databases, uncertainty in RDF knowledge graphs and ontologies for sensor and numeric data.

\section{Related Work}
\label{sec:relatedwork}

\paragraph{Similarity-Aware Extensions of SPARQL.}
Several works extend SPARQL with similarity mechanisms over
strings~\cite{DBLP:conf/semweb/KieferBS07}, scalar
attributes~\cite{DBLP:conf/semweb/FerradaBH20,DBLP:journals/semweb/FerradaBH24},
nearest-neighbour joins~\cite{DBLP:journals/pacmmod/ArroyueloBGHNR24}, or
embedding vectors~\cite{DBLP:conf/swat4ls/KulmanovKKMGDH18}. These approaches
make approximate matching available for deterministic objects, whereas
ProbSPARQL targets uncertain numeric measurements whose values are probability
distributions. The resulting query problem is therefore distributional
compatibility between uncertain measurements, rather than approximate matching
between deterministic values.

\paragraph{Probabilistic Relational Databases.}
A foundational line of probabilistic database work reasons about tuple-level
uncertainty and query answers through possible-world or lineage-based
semantics~\cite{DBLP:journals/cacm/DalviRS09,DBLP:conf/vldb/AgrawalBSHNSW06,DBLP:conf/sigmod/HuangAKO09}.
Other systems support attribute-level uncertainty or simulation-based query
evaluation~\cite{DBLP:conf/sigmod/SinghMMPHS08,DBLP:conf/sigmod/JampaniXWPJH08},
and interval- or bound-based approaches improve tractability for complex
queries~\cite{DBLP:conf/sigmod/FengGHK21}. These systems provide foundations
for uncertain data management in relational schemas, whereas ProbSPARQL targets
RDF graphs where uncertain numeric measurements must remain connected to
component identity, units, provenance, and other KG context.

\paragraph{Uncertainty in RDF Knowledge Graphs.}
A major line of work on uncertainty in RDF and knowledge graphs attaches
uncertainty to triples or annotations \cite{DBLP:conf/semweb/Fukushige05,DBLP:conf/iri/UdreaSM06,fang2016psparql,DBLP:journals/ws/ZimmermannLPS12,DBLP:conf/rr/Straccia09,DBLP:conf/emnlp/ZhuWWZCKS25},
while probabilistic ontology formalisms attach uncertainty to axioms or derived
entailments~\cite{DBLP:conf/semweb/CostaLL08,DBLP:conf/aaai/KollerLP97,DBLP:conf/semweb/ZhuPXTNSK23,DBLP:conf/jelia/GiugnoL02,DBLP:conf/uai/ZhuPXTNKS26}.
These frameworks primarily model uncertainty about the truth, confidence, or
degree of graph statements. ProbSPARQL addresses a different layer in which graph statements remain deterministic, while uncertain numeric measurement values are queried as probability distributions within SPARQL query evaluation~\cite{DBLP:journals/corr/abs-2605-16568}.

\paragraph{Ontologies for Sensor and Numeric Data.}
Vocabularies such as ProbOnto~\cite{swat2016probonto}, SSN~\cite{w3c-ssn},
and the Data Cube~\cite{W3C:DataCube} describe probability distributions,
sensor observations, provenance, and statistical datasets in RDF. These
vocabularies provide a descriptive layer, but not executable SPARQL semantics for
querying and combining distribution-valued measurements. ProbSPARQL complements
them with query-layer operations over distribution-valued probabilistic literals.

\section{Discussion, Impact, and Conclusion}

\paragraph{Practical gap.}
ProbSPARQL addresses the gap between uncertain numeric measurements produced by
circular-factory inspection and manufacturing workflows and the deterministic literal model of
current RDF/SPARQL systems. It keeps measurement uncertainty connected to the
surrounding product, component, unit, and measurement context, while making
probabilistic filtering, propagation, comparison, and matching executable inside
the SPARQL query layer.

\paragraph{Integration impact.}
The main integration benefit is that uncertainty-aware query operations remain
inside the KG workflow rather than a separate application-layer pipeline. An ordinary SPARQL pipeline with external numerical processing must
retrieve candidate distributions before filtering or matching can be applied.
Expanding distribution parameters into RDF triples keeps the data inside the
graph, but it makes query patterns verbose and shifts datatype-specific
operations into query logic or external code. A separate probabilistic data store
can support numerical processing, but it separates uncertainty handling from the
product, component, unit, and measurement context. ProbSPARQL keeps these
operations within the same KG interface for downstream uncertainty-aware access.

\paragraph{Pilot status.}
The prototype is an active query-layer pilot being integrated into the SFB~1574
circular-factory infrastructure, not yet a production service. Current and
scheduled take-up spans the data-infrastructure, inspection, material-analysis,
planning/control, reliability-modeling, and reassembly-planning teams. The pilot
validates the modeling pattern and partner-reviewed query templates on
project-derived measurement fragments. The evaluation demonstrates feasible
in-engine execution, effective filter pushdown, configurable divergence-join
strategies, and datatype extensibility across three probabilistic representations.

\paragraph{Lessons and limitations.}
The pilot yields three lessons.
Distribution-valued literals need explicit random-variable nodes to preserve
domain, unit, and measurement context. Selective probabilistic filters benefit
from in-engine evaluation by avoiding candidate-distribution transfer.
Divergence-based matching needs configurable decision strategies for
module-specific latency and accuracy requirements. Remaining challenges include
worst-case quadratic candidate enumeration for divergence joins,
compatible-domain mappings for cross-datatype comparison, and validation with
downstream engineering modules. Future work will investigate index-supported
pruning, richer unit and domain checks, and feedback from broader pilot
deployment in circular-factory~workflows.

\section*{Supplemental Material Statement}
The supplement~\cite{miscprobsparql2026} provides detailed formal semantics, complete query workloads, extended evaluation protocols and results, algebraic properties and proofs, and implementation and reproducibility details. The implementation, generated evaluation data, query workloads, and replication scripts introduced above are additionally archived with a persistent identifier on DaRUS~\cite{darusprobsparql2026}.

\section*{Acknowledgments}
The project was funded by the Deutsche Forschungsgemeinschaft (DFG, German Research
Foundation) - SFB 1574 - Project number 471687386 (``Circular Factory''). The
authors thank the International Max Planck Research School for Intelligent Systems
(IMPRS-IS) for supporting Jingcheng Wu and Hongkuan~Zhou. The authors also thank
Jianyang Gu for reviewing the code.

\section*{Declaration on the Use of Generative AI}
ChatGPT was used to assist with language editing of this Work, including grammar, wording, clarity, and stylistic refinement, and to provide suggestions for improving figure readability. All AI-assisted edits and suggestions were reviewed and verified by the authors, who remain fully responsible for the content.

\bibliographystyle{splncs04}
\bibliography{references}

\appendix
\newcommand{\includedfrommain}{}
% \subfile{supplement.tex}

\end{document}